\documentclass{article}

\PassOptionsToPackage{numbers,compress,sort}{natbib}

\usepackage[main,final]{neurips_2026}

\usepackage[utf8]{inputenc} 
\usepackage[T1]{fontenc}    
\usepackage{hyperref}       
\usepackage{url}            
\usepackage{booktabs}       
\usepackage{amsfonts}       
\usepackage{amsmath}        
\usepackage{amssymb}        
\usepackage{amsthm}         
\usepackage{mathrsfs}       
\usepackage{graphicx}       
\usepackage{nicefrac}       
\usepackage{microtype}      
\usepackage{xcolor}         

\newtheorem{theorem}{Theorem}

\title{Understanding Double Descent through Universal Compression}
\hypersetup{
  pdftitle={Understanding Double Descent through Universal Compression},
  pdfauthor={Jiaxuan Cheng and Addison Spiegel}
}

\author{%
  Jiaxuan Cheng\\
  Massachusetts Institute of Technology\\
  \texttt{jasonc75@mit.edu}
  \And
  Addison Spiegel\\
  Massachusetts Institute of Technology\\
  \texttt{spiegel1@mit.edu}
}

\begin{document}

\maketitle

\begin{abstract}
  Many deep learning models exhibit a \emph{double-descent} phenomenon,
  where test loss exhibits a sharp peak near the interpolation threshold and
  then descends a second time as model size grows further, contrary to the
  U-shape predicted by the classical bias--variance tradeoff. The
  double-descent phenomenon has been found and explained in toy classical
  models, but is not well-understood in modern deep learning despite
  extensive empirical study. We interpret double
  descent through the novel lens of universal compression. Specifically, we
  upper bound the maximum-likelihood excess test loss as a sum of four terms: an
  approximation-and-shift term, a minimax batch regret, a prior cost, and an
  MLE-vs-Bayes gap. Under exact posterior sampling, we further show that the expected
  MLE-vs-ensemble gap lower-bounds the MLE-vs-Bayes gap in the well-specified
  setting. In our experiments, a Gaussian-prior SGLD approximation
  to the Bayes predictor, unlike SGD, shows no pronounced interpolation
  peak. Experiments with uniform random labels support this gap as a
  major contributor to the double-descent peak; here, the SGD cross-entropy is
  roughly four times the SGLD ensemble cross-entropy at the peak, while
  the ensemble cross-entropy remains near $\log 10$.

\end{abstract}


\section{Introduction}
\label{sec:intro}

The test loss of a neural network, plotted against model size, does not behave
the way classical bias--variance reasoning predicts. Near the interpolation
threshold the test loss rises into a sharp peak and then descends a second
time, a phenomenon known as \emph{double descent}
\citep{belkin2019reconciling,nakkiran2021deep}. Despite extensive empirical
study, it remains unclear which component of the test loss is responsible for
the peak, and why that component then recovers as the model class grows
further.

The reason why double descent occurs is still not well understood in full
generality. Several explanations have been proposed:
\citet{belkin2019reconciling} attribute the phenomenon to an implicit bias of
overparameterized models toward simpler interpolating solutions, and
\citet{nakkiran2021deep} introduce \emph{effective model complexity} as a
unifying scale and observe that the peak occurs when this complexity is
roughly equal to the sample size. These accounts are intuitive but remain
descriptive, and are difficult to rigorously prove or directly test.

We approach the question through the lens of universal compression. We give
an upper bound on the excess test cross-entropy of the trained model that decomposes
into four interpretable terms: an \emph{approximation-and-shift term},
comparing plug-in risk under the true distribution and its in-class projection;
a \emph{minimax batch regret}, the unavoidable worst-case cost any predictor
incurs to encode data drawn from the class without knowing its parameters;
a \emph{prior cost}, measuring the
excess worst-case risk of the chosen Bayes predictor over the minimax value;
and an \emph{MLE-vs-Bayes gap}, the extra cost of using the trained
maximum-likelihood model in place of the Bayes predictor.

We test this picture on width-varied CNNs trained on CIFAR-10 with some or
all of the training labels replaced by uniform noise. We approximate a
Bayes predictor by averaging predictive class probabilities from
$K$ independently initialized SGLD chains under an explicit Gaussian prior.
Under 20\% label noise the per-seed test cross-entropy exhibits the canonical
double-descent peak while the SGLD ensemble does not. To eliminate the
approximation-and-shift term, we additionally run experiments under fully
random labels, where this term is zero by construction; in this setting the
SGD peak persists, empirically supporting the MLE-vs-Bayes gap as a major
contributor to the peak.

Our contributions can be summarized as follows: (1) We bound the MLE excess test
cross-entropy by a sum of four terms, motivated by universal compression.
(2) Under exact posterior sampling, we lower-bound the MLE-vs-Bayes gap
by the expected MLE-vs-ensemble loss difference in the well-specified
setting, and support this attribution experimentally under fully random
labels and an explicit Gaussian prior.

\begin{figure}[t]
  \centering
  \includegraphics[width=\linewidth]{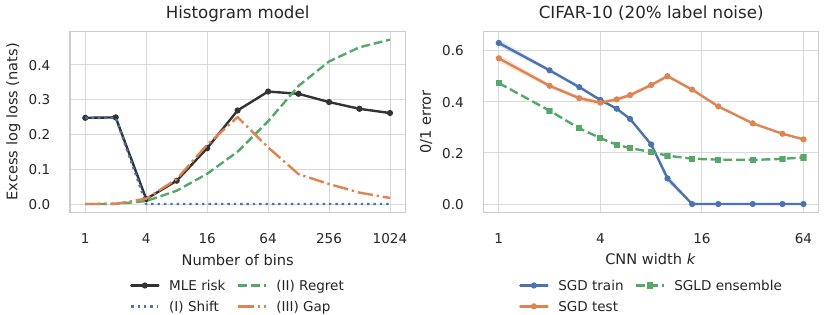}
  \caption{\textbf{Right:} 5-layer CNNs of varying width $k$ trained on
  CIFAR-10 with 20\% label noise ($n=50{,}000$, SGD, $200$k steps). The
  per-seed classification error (orange) exhibits the double-descent peak near
  the interpolation threshold, while the probability average over 8 SGLD
  chains (green, dashed) reduces the peak.
  \textbf{Left:} A histogram model with $n=64$ uniform inputs on $[0,1]$,
  adapted from Holzm\"uller (2021, Appendix~D). The class probabilities on
  four equal intervals are $(.05,.95,.5,.5)$, and each fitted bin chooses
  from $\{.05,.5,.95\}$, using $.5$ if empty. Finite sums give the MLE excess
  loss and terms~(I)--(III). Once $d\ge4$, term~(I) is zero, batch
  regret~(II) increases with $d$, and the MLE-vs-Bayes gap~(III) exhibits a peak.
  Here the prior is least-favorable, so (II') is zero.}
  \label{fig:ensemble-collapse}
\end{figure}


\section{Related work}
\label{sec:related}

\paragraph{Double descent in deep learning.}
\citet{belkin2019reconciling,nakkiran2021deep} first articulated the
double-descent picture across kernel methods, small neural networks, and
modern deep networks (CNNs, ResNets), showing that the peak in test loss
arises along several axes simultaneously---including model width, training
time, and dataset size---and is amplified by label noise. Subsequent work has refined the picture: sample-wise variants
\citep{nakkiran2020optimal}, the role of regularization in attenuating the
peak \citep{nakkiran2021optimal}, empirical characterizations of when the
phenomenon appears \citep{schaeffer2023double}, and the related grokking
phenomenon \citep{power2022grokking}.
Finally, \citet{wilson2020bayesian} observe empirically that Bayesian
model averaging (approximated by deep ensembles) can eliminate the
double-descent peak. We study this observation by decomposing the MLE
excess test cross-entropy into four terms through the universal-compression
lens---an approximation-and-shift term, a minimax batch regret, a prior cost,
and an MLE-vs-Bayes gap---and, under $100\%$ label noise where the
approximation-and-shift term is zero by construction, find a large
SGD-vs-ensemble gap near interpolation.

\paragraph{Double descent in classical models.}
A complementary line of work analyzes double descent in tractable, mostly
linear, settings: high-dimensional linear regression
\citep{advani2017high,hastie2022surprises}, finite-sample least squares on
concrete data models \citep{belkin2020two}, benign overfitting of
minimum-norm interpolants \citep{bartlett2020benign,muthukumar2020harmless},
ridge regression over random features \citep{mei2022generalization}, and
feature designs exhibiting arbitrarily many descents \citep{chen2021multiple}.
Our contribution is complementary: we target \emph{misspecified, non-regular}
model families, of which deep networks are the prototypical case, rather
than well-specified linear or kernel settings. Unlike these exact risk
analyses, our bound separates candidate contributions without giving a
tight risk formula for deep networks.

\paragraph{Information-theoretic complexity and deep learning theory.}
For discrete data, $-\log q(x)$ is an ideal code length in nats, so expected
excess code length over the true source is a KL divergence.
Classical complexity measures such as VC dimension \citep{vapnik1971uniform}
and Rademacher complexity \citep{bartlett2002rademacher} are notoriously
loose for overparameterized networks; mutual-information generalization
bounds \citep{xu2017information,steinke2020reasoning} provide a
distribution-dependent alternative. Universal coding
\citep{rissanen1984universal,barron1998minimum,davisson1973universal,shtarkov1987universal,haussler1997mutual,merhav1998universal,rissanen1996fisher}
characterizes the unavoidable cost of predicting without knowing the
source parameter, and singular learning theory
\citep{watanabe2009algebraic,hoogland2024towards,chen2023dynamical}
extends this to the non-regular families characteristic of deep networks.
However, none of these frameworks have been shown to directly explain the
double-descent phenomenon in full generality.


\section{Decomposition of the MLE test loss}
\label{sec:methods}

This section develops the analytic object the rest of the paper studies. We fix
notation and write the test loss of the maximum-likelihood plug-in as a single KL
quantity (Section~\ref{sec:setup}), introduce reference quantities that will
appear in the decomposition (Section~\ref{sec:refobj}), and state the inequality
that decomposes test loss into approximation, batch-regret, prior-cost, and MLE-vs-Bayes
contributions (Section~\ref{sec:decomp}). Section~\ref{sec:gibbs} describes how
we approximate a Bayes predictor in practice using SGLD.
Appendix~\ref{app:interpretation} summarizes the notation and the quantities
measured in the experiments.

\subsection{Setup}
\label{sec:setup}

We consider a measurable space $(\mathcal{X},\mu)$ and a true data-generating
distribution $P^*\in\mathcal{P}(\mathcal{X})$ with density $p^*=dP^*/d\mu$.
We draw training data $X_1,\dots,X_n$ and a fresh test point $X_{n+1}$ i.i.d.\
from $P^*$. A model class is a family
$\{P_{X\mid\theta}:\theta\in\Theta\}$ on $(\mathcal{X},\mu)$ with densities
$p_\theta$, which we extend to product form
$P_{X^n\mid\theta}=\prod_{i=1}^n P_{X\mid\theta}$. We make no regularity,
identifiability, smoothness, or finite-dimensionality assumption on $\Theta$, and
we allow $P^*$ to lie outside the class. We take the maximum-likelihood
estimator
\begin{equation*}
\hat\theta_n=\hat\theta_n(X^n)\in\arg\max_{\theta\in\Theta}\prod_{i=1}^n p_\theta(X_i)
\end{equation*}
to be a measurable selection, breaking ties arbitrarily. We assume the
entropies, risks, and differences below are finite. The same decomposition holds for any
measurable plug-in rule, with expectations also taken over its internal
randomness when applicable.

\paragraph{Test loss.}
We define the MLE plug-in predictor as the density
$\hat p_n\triangleq p_{\hat\theta_n(X^n)}$ and write its test log-loss as
\begin{equation}
L_n(P^*) \;\triangleq\; \mathbb{E}_{X^n,X_{n+1}\stackrel{\textit{i.i.d.}}{\sim}P^*}\!\big[\log 1/\hat p_n(X_{n+1})\big].
\label{eq:testloss}
\end{equation}
For other predictive rules $Q$, we write $L^Q_n(P)$ for the analogous loss.
Subtracting the entropy $H(P^*)\triangleq-\mathbb{E}_{P^*}[\log p^*(X)]$ and applying
the chain rule of KL divergence gives the identity
\begin{equation}
\mathcal{R}_n(P) \;\triangleq\; L_n(P)-H(P)
\;=\; \mathbb{E}_{X^n\sim P^{\otimes n}}\!\Big[D\!\left(P\,\big\|\,p_{\hat\theta_n(X^n)}\right)\Big],
\quad P\in\mathcal{P}(\mathcal{X}),
\label{eq:plugin-risk}
\end{equation}
which we call the \emph{plug-in risk} of MLE under $P$. We define
$\mathcal{R}_n(\cdot)$ for every $P$, in or out of the class; it depends on
$\Theta$ only through the MLE.

\paragraph{Supervised learning.}
Our setup generalizes immediately to supervised data, by reducing a supervised
problem to an unsupervised one with a fixed input marginal. Formally:
for $(X_i,Y_i)\stackrel{\textit{i.i.d.}}{\sim}P^*_{XY}$, we fix the
$X$-marginal $P_X=P_X^*$ and let the model
parameterize the joint
$P_{XY\mid\theta}(x,y)=P_X(x)\,P_{Y\mid X,\theta}(y\mid x)$. Substituting
$\tilde X_i\triangleq(X_i,Y_i)$ recovers the present setup; the constant
$H(P_X)$ cancels from both sides of the decomposition stated below; and the
resulting MLE reduces to the conditional MLE
$\hat\theta_n\in\arg\max_\theta\prod_i P_{Y\mid X,\theta}(Y_i\mid X_i)$.
For this reason, we develop the rest of the theory in the unsupervised setup
with a single variable $X\sim P^*$ rather than separate $X$ and $Y$, without
loss of generality.

\subsection{Reference predictors}
\label{sec:refobj}

The following quantities indexed by the model class and prior will enter the decomposition.

\paragraph{In-class KL projection.}
We define the closest in-class density to $P^*$ in the KL sense as
\begin{equation}
P^*_\Theta \;\triangleq\; P_{X\mid\theta^*(P^*)},
\qquad
\theta^*(P^*) \in \arg\!\inf_{\theta\in\Theta} D\!\left(P^*\,\big\|\,P_{X\mid\theta}\right).
\label{eq:projection}
\end{equation}
We assume the infimum in~\eqref{eq:projection} is attained.
Note $P^*=P^*_\Theta$ exactly when
$P^*$ lies in the class.

\paragraph{Bayes predictors and batch regret.}
Following \citet{polyanskiy-wu}, we define the i.i.d.\ batch redundancy
\begin{equation}
\mathrm{BatchReg}^*_{n+1}(\Theta) \;\triangleq\;
\inf_{Q_{n+1}(\,\cdot\,;\,x^n)} \,\sup_{\theta_0\in\Theta}\,
\mathbb{E}_{X^n\sim P_{X^n\mid\theta_0}}\!\Big[D\!\left(P_{X\mid\theta_0}\,\big\|\,Q_{n+1}(\,\cdot\,;\,X^n)\right)\Big].
\label{eq:batchreg}
\end{equation}
A \emph{minimax predictor} $Q^*_{n+1}$ is any minimizer of the outer infimum
(when attained). For a proper
prior $\pi$ with a well-defined posterior, the Bayes predictor is
\begin{equation}
Q^\pi_{n+1}(x_{n+1};\,x^n)
= \mathbb{E}_{\theta\sim\pi(\,\cdot\mid X^n=x^n)}
   \!\big[p_\theta(x_{n+1})\big].
\label{eq:bayesmix}
\end{equation}
Write its worst-case in-class risk and its \emph{prior cost} as
\begin{align}
\mathrm{BatchReg}^\pi_{n+1}(\Theta)
&\triangleq \sup_{\theta_0\in\Theta}
\mathbb{E}_{X^n\sim P_{X^n\mid\theta_0}}
\!\Big[D\!\left(P_{X\mid\theta_0}\,\big\|\,Q^\pi_{n+1}(\,\cdot\,;\,X^n)\right)\Big],
\label{eq:prior-risk}\\
\Gamma^\pi_{n+1}(\Theta)
&\triangleq \mathrm{BatchReg}^\pi_{n+1}(\Theta)
-\mathrm{BatchReg}^*_{n+1}(\Theta)\ge 0.
\label{eq:prior-cost}
\end{align}
Non-negativity follows because $Q^\pi_{n+1}$ is feasible in~\eqref{eq:batchreg}.
When a least-favorable prior $\pi^*$ exists whose Bayes predictor attains
the minimax value, $\Gamma^{\pi^*}_{n+1}(\Theta)=0$.

\paragraph{MLE-vs-Bayes gap.}
For $\theta_0\in\Theta$, define
\begin{equation}
\Delta^{\mathrm{MLE},\pi}_n(\theta_0)
\;\triangleq\;
\mathcal{R}_n(P_{X\mid\theta_0})
\;-\;
\mathbb{E}_{X^n\sim P_{X^n\mid\theta_0}}\!\Big[D\!\left(P_{X\mid\theta_0}\,\big\|\,Q^\pi_{n+1}(\,\cdot\,;\,X^n)\right)\Big],
\label{eq:DeltaMLE}
\end{equation}
the difference at the well-specified $\theta_0$ between the plug-in risk of MLE
and the in-class risk of $Q^\pi_{n+1}$.

\subsection{Decomposition}
\label{sec:decomp}

The plug-in risk under $P^*$ admits the following bound.

\begin{theorem}[Test-loss decomposition]
\label{thm:decomp}
For every proper prior $\pi$ satisfying the assumptions above and every $n\ge 1$,
\begin{equation}
\begin{aligned}
L_n(P^*)-H(P^*)
&\;\le\;
\underbrace{\mathcal{R}_n(P^*)-\mathcal{R}_n(P^*_\Theta)}_{\text{(I)\ approximation \& shift}}
\;+\;
\underbrace{\mathrm{BatchReg}^*_{n+1}(\Theta)}_{\text{(II)}}\\
&\quad+\;
\underbrace{\Gamma^\pi_{n+1}(\Theta)}_{\text{(II') prior cost}}
\;+\;
\underbrace{\sup_{\theta_0\in\Theta}\Delta^{\mathrm{MLE},\pi}_n(\theta_0)}_{\text{(III)\ MLE-vs-Bayes gap}}.
\end{aligned}
\label{eq:decomp}
\end{equation}
\end{theorem}
\begin{proof}
By~\eqref{eq:plugin-risk}, $L_n(P^*)-H(P^*)=\mathcal{R}_n(P^*)$. Adding and
subtracting $\mathcal{R}_n(P^*_\Theta)$ gives
\begin{equation*}
L_n(P^*)-H(P^*)
\;=\;\big[\mathcal{R}_n(P^*)-\mathcal{R}_n(P^*_\Theta)\big]
\;+\;\mathcal{R}_n(P^*_\Theta).
\end{equation*}
Since $P^*_\Theta=P_{X\mid\theta^*(P^*)}$ with $\theta^*(P^*)\in\Theta$, we have
$\mathcal{R}_n(P^*_\Theta)\le\sup_{\theta_0\in\Theta}\mathcal{R}_n(P_{X\mid\theta_0})$.
Decomposing the well-specified plug-in risk via~\eqref{eq:DeltaMLE},
\begin{equation*}
\mathcal{R}_n(P_{X\mid\theta_0})
\;=\;
\mathbb{E}_{X^n\sim P_{X^n\mid\theta_0}}\!\Big[D\!\left(P_{X\mid\theta_0}\,\big\|\,Q^\pi_{n+1}(\,\cdot\,;\,X^n)\right)\Big]
\;+\;\Delta^{\mathrm{MLE},\pi}_n(\theta_0),
\end{equation*}
and applying $\sup(a+b)\le\sup a+\sup b$ together with the definition
of $\mathrm{BatchReg}^\pi_{n+1}(\Theta)$ in~\eqref{eq:prior-risk} yields
(II) + (II') + (III).
\end{proof}

We use the right-hand side to separate candidate contributions to test loss. We now
describe each term and how it relates to double descent.

\paragraph{(I) Approximation and shift.}
The first term is the difference between the plug-in risk of MLE on the experiment
generated by $P^*$ and on the experiment generated by its in-class projection
$P^*_\Theta$. We note that this is not the textbook approximation error
$D(P^*\|P^*_\Theta)$. It need not be non-negative or monotone at finite $n$,
but vanishes when $P^*=P^*_\Theta$.

\paragraph{(II) Batch regret.}
The second term is the textbook minimax batch regret of $\Theta$. It is
monotonically non-decreasing as $\Theta$ grows: for $\Theta_1\subseteq\Theta_2$,
any $Q$ feasible in the outer infimum of~\eqref{eq:batchreg} for $\Theta_2$ is
also feasible for $\Theta_1$, while the inner supremum runs over a larger set,
so $\mathrm{BatchReg}^*_{n+1}(\Theta_1)\le\mathrm{BatchReg}^*_{n+1}(\Theta_2)$.
The additional prior cost (II') need not be monotone across model classes.

\paragraph{(III) MLE-vs-Bayes gap.}
The MLE-vs-Bayes gap does not admit a clean
asymptotic description in the non-regular settings characteristic of
deep neural networks. Our experiments find a large SGD-vs-ensemble gap
near the interpolation threshold, consistent with a large contribution
from~(III). We depict the
intuition behind this in the left panel of Figure~\ref{fig:ensemble-collapse}.

\subsection{Approximating $Q^\pi_{n+1}$ by a Gibbs ensemble}
\label{sec:gibbs}

The Bayes predictor $Q^\pi_{n+1}$ in~\eqref{eq:bayesmix} is rarely available in closed
form. However, we can approximate it by Monte~Carlo. For an inverse temperature $\beta>0$
and a prior $\pi$ on $\Theta$, the \emph{log-loss Gibbs estimator} is
\begin{equation}
\rho_{\beta,\pi}(d\theta\mid x^n)
\;\propto\; \exp\!\big\{-\beta\,\mathcal{L}_n(\theta;x^n)\big\}\,\pi(d\theta),
\qquad
\mathcal{L}_n(\theta;x^n)\triangleq \sum_{i=1}^n\log\tfrac{1}{p_\theta(x_i)},
\label{eq:gibbs}
\end{equation}
and at $\beta=1$, $\rho_{1,\pi}(\,\cdot\mid x^n)$ is exactly the Bayes posterior
under $\pi$.

For continuous parameters, we use temperature-one SGLD with unbiased
stochastic gradients of the mean negative log-likelihood and the prior drift in
Appendix~\ref{app:training}. Take
\[
\eta_t=\eta_0(1+t/\tau)^{-\alpha},\qquad
\tfrac12<\alpha\le1,\qquad
\operatorname{Cov}(\varepsilon_t)=2\eta_t I/n,
\]
where $\eta_0,\tau>0$ and $\varepsilon_t$ is the injected Gaussian noise.
Under the stability assumptions of \citet[Theorem~7]{teh2016sgld},
the step-size-weighted empirical distribution converges weakly to the posterior
as the number of steps tends to infinity.
For the sampling results below, we assume an ideal oracle that returns
independent draws from this limiting posterior.

Drawing
$\theta^{(1)},\dots,\theta^{(K)}\stackrel{\textit{i.i.d.}}{\sim}\rho_{1,\pi}(\,\cdot\mid X^n)$
and averaging the per-sample predictives gives the posterior-predictive Monte~Carlo
estimator
\begin{equation}
\widehat Q^{(K)}_{n+1}(\,\cdot\,;\,X^n)
\;\triangleq\; \frac{1}{K}\sum_{k=1}^{K} p_{\theta^{(k)}}(\cdot)
\;\xrightarrow[K\to\infty]{\text{a.s.}}\;
Q^\pi_{n+1}(\,\cdot\,;\,X^n),
\label{eq:gibbs-mc}
\end{equation}
pointwise in $X^n$ and in the test argument \citep[Ch.~6]{robert2007bayesian}.

\paragraph{Lower-bounding the MLE-vs-Bayes gap.}
Under the posterior-sampling assumption, the ensemble is pointwise
unbiased for $Q^\pi_{n+1}$ conditional on $X^n$.
Jensen's inequality applied to \eqref{eq:gibbs-mc} gives
\begin{equation}
\mathbb{E}_{\theta^{(1)},\ldots,\theta^{(K)}}\!\left[L^{\widehat Q^{(K)}}_n(P^*)\right]
\;\ge\; L^{Q^\pi}_n(P^*),
\label{eq:jensen-bound}
\end{equation}
so an unbiased predictive approximation has test log-loss at least that of
$Q^\pi_{n+1}$ in expectation over posterior sampling. Setting $\theta_0=\theta^*(P^*)$
in~\eqref{eq:DeltaMLE} under $P^*_\Theta=P^*$ gives
$\Delta^{\mathrm{MLE},\pi}_n(\theta^*(P^*))=L_n(P^*)-L^{Q^\pi}_n(P^*)$;
combining with~\eqref{eq:jensen-bound},
\begin{equation}
L_n(P^*) - \mathbb{E}_{\mathrm{MC}}\!\left[L^{\widehat Q^{(K)}}_n(P^*)\right]
\;\le\; \Delta^{\mathrm{MLE},\pi}_n(\theta^*(P^*))
\;\le\; \sup_{\theta_0\in\Theta}\Delta^{\mathrm{MLE},\pi}_n(\theta_0)
\;=\; (\mathrm{III}).
\label{eq:gap-lower-bounds-III}
\end{equation}
Thus, in the well-specified setting, the expected MLE-vs-ensemble gap
lower-bounds (III) for every ensemble size $K$ under the posterior-sampling
assumption.


\section{Experiments}
\label{sec:experiments}

\begin{figure}[t]
  \centering
  \includegraphics[width=\linewidth]{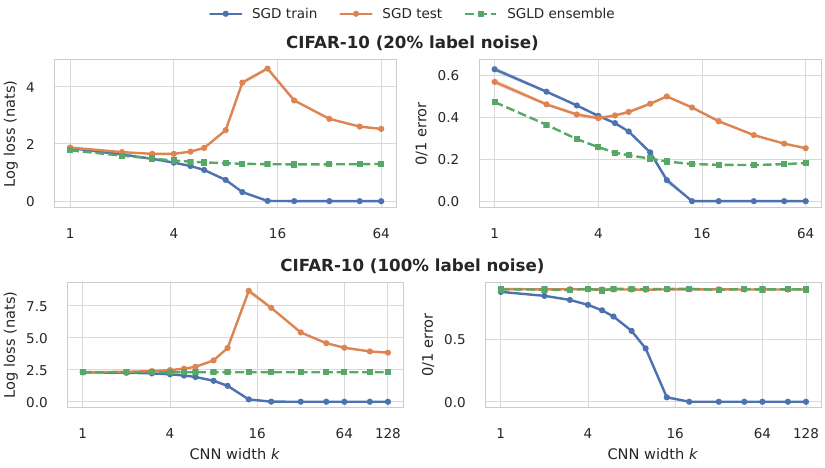}
  \caption{CIFAR-10 with 20\% label noise (top) and 100\% label noise
  (bottom), with $K=8$ SGD seeds per width. We plot cross-entropy of the train set, the per-seed
  test set (mean over seeds; band $\pm 1$ s.d.), and the probability-averaged
  ensemble of 8 Gaussian-prior SGLD chains with 10 retained draws per chain
  (green, dashed).
  Under both noise levels the per-seed test cross-entropy exhibits a
  pronounced double-descent peak near the interpolation threshold while the ensemble
  cross-entropy is smaller, particularly near interpolation.
  Test cross-entropy uses independently corrupted labels and 0/1 error uses clean labels.
  We include 0/1 error for reference, but note that it cannot separate near-uniform
  predictions from overconfident wrong predictions in the 100\% noise case---the
  distinction cross-entropy resolves.}
  \label{fig:ensemble}
\end{figure}

We investigate this picture across two model families that span the standard
double-descent literature---CNNs and random forests---each with a varying
parameter count across the interpolation threshold on a classification
task. Throughout we report \emph{cross-entropy} as the
primary metric. Cross-entropy is the natural quantity in our framework because
it estimates the test log-loss $L_n(P^*)$ in~\eqref{eq:testloss} and therefore
measures predictive density rather than 0/1 correctness; classification accuracy
is reported for context but does not enter the decomposition~\eqref{eq:decomp}.

\paragraph{Ensemble as a Bayes predictor.}
Following Section~\ref{sec:gibbs}, we approximate the Bayes predictor
$Q^\pi_{n+1}$ using finite runs of $K$ independent SGLD chains under an explicit Gaussian
prior, averaging retained predictive probabilities within each chain and
then across chains. Crucially, we
average \emph{class probabilities} (the softmax outputs), \emph{not logits}.

\begin{figure}[t]
  \centering
  \includegraphics[width=\linewidth]{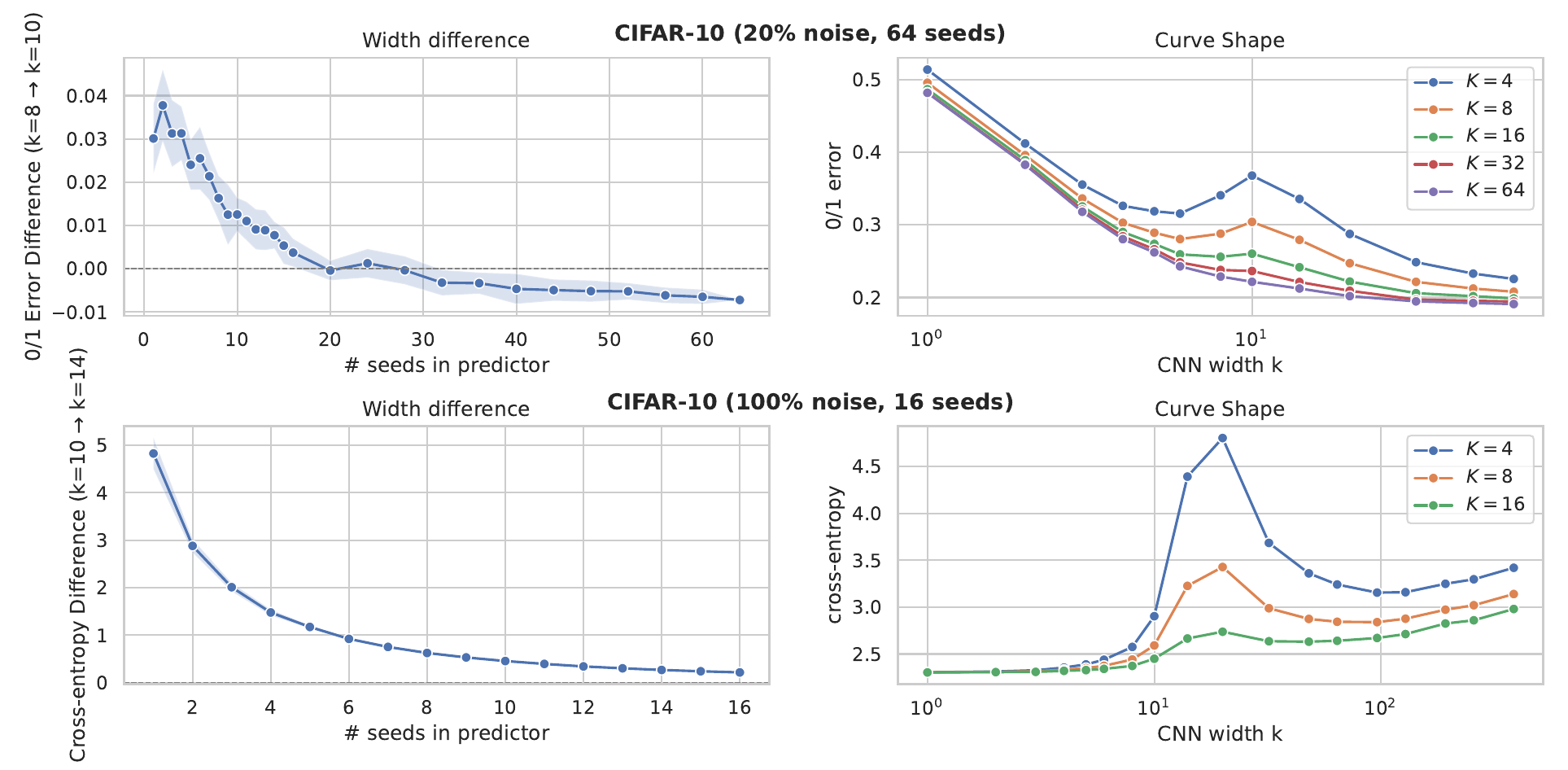}
  \caption{Effect of the number of seeds $K$ averaged into the SGD ensemble.
  Left panels: the difference in the ensemble test metric between two widths
  ($k{=}8\to 10$ for 20\% noise, $k{=}10\to 14$ for 100\% noise),
  computed as the larger width minus the smaller, as a function of $K$.
  Right panels: the ensemble test curve as $K$ varies. We report 0/1 error
  for 20\% noise and cross-entropy for 100\% noise. Lines give means over
  20 random subsets of 64 seeds (20\% noise) or 16 seeds (100\% noise),
  with no replacement within each subset. Left-panel subsets are paired
  across widths; bands give $\pm 1$ s.d. The full ensemble is used once.
  Under 20\% noise the difference is negative for $K\ge32$.
  Under 100\% noise, the difference remains positive at $K=16$.}
  \label{fig:gap-vs-seeds}
\end{figure}

\subsection{Bayes predictor and double descent}
\label{sec:exp-ensemble}

We train a 5-layer CNN of varying width $k$ on CIFAR-10 with a fixed
fraction of training labels replaced by uniformly random labels, sweeping widths. For each
$k$ we train $K=8$ independently initialized SGD seeds and $K=8$ SGLD
chains, and compare the mean per-seed SGD test cross-entropy with the
SGLD ensemble test cross-entropy. SGD is a finite optimization of the
maximum-likelihood objective. Full
training details (architecture, optimizer, schedule, batch size, number
of gradient steps) are deferred to Appendix~\ref{app:training}.

\paragraph{CIFAR-10.}
We focus first on the canonical $20\%$ label noise setup
(Figure~\ref{fig:ensemble}). The training cross-entropy approaches
zero at the interpolation threshold and remains close to zero for all larger $k$,
confirming that we are operating in the interpolation regime. The per-seed
test cross-entropy exhibits the double-descent peak near the interpolation
threshold, while the SGLD ensemble cross-entropy
shows no pronounced interpolation peak. We hypothesize that
the second descent is primarily produced by the MLE-vs-Bayes gap~(III),
but cannot conclude this directly due to the possible confounding effect
of the approximation-and-shift term~(I).

\paragraph{Random labels.}
To isolate the MLE-vs-Bayes gap~(III) from the approximation-and-shift
term~(I), we repeat the experiment with $100\%$ label noise: training labels
drawn uniformly at random, independently of the input. Under this $P^*$ the
true conditional is uniform, $P^*(Y\mid X)\equiv 1/10$. Setting the final-layer weights and biases to zero gives uniform
softmax outputs, so there is a $\theta_0\in\Theta$ with
$D(P^*\,\|\,P_{\theta_0})=0$. Hence
$P^*_\Theta=P^*$ in distribution and
$(\mathrm{I})=\mathcal{R}_n(P^*)-\mathcal{R}_n(P^*_\Theta)=0$.

The training cross-entropy approaches zero at the interpolation threshold, and
the per-seed test cross-entropy exhibits a pronounced peak there
(Figure~\ref{fig:ensemble}). The ensemble test cross-entropy remains near
$\log 10$, so the empirical SGD-vs-ensemble gap is large near the
interpolation threshold. At $k=14$, the mean SGD loss is $8.659$ nats and
the SGLD ensemble loss is $2.303$ nats, giving a gap of $6.356$ nats.
Since $P^*_\Theta=P^*$ here by construction,
\eqref{eq:gap-lower-bounds-III} identifies the expected MLE-vs-ensemble gap
as a lower bound on (III) in the ideal posterior-sampling model.
The large measured gap near interpolation supports the MLE-vs-Bayes gap
as a major contributor to the peak.

Longer chains and a smaller step size changed ensemble log-loss by at most
$0.019$ nats at the checked widths (Appendix~\ref{app:sampling}).
These checks preserve the qualitative comparison. An SGD-ensemble
control is also included there.

\paragraph{SGD ensembling.}
As a practical extension, we also average the class probabilities of
independently trained SGD models.
Figure~\ref{fig:gap-vs-seeds} compares the ensemble curves as $K$ varies.
Increasing $K$ improves the ensemble test metric. Under $20\%$ noise,
the plotted ensemble curves are monotone for $K=32$ and $K=64$; under
$100\%$ noise, the peak shrinks but remains at $K=16$.

\begin{figure}[t]
  \centering
  \includegraphics[width=\linewidth]{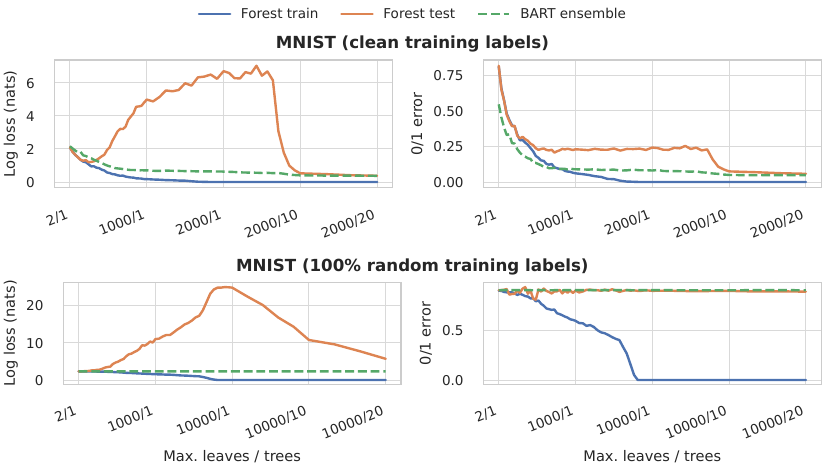}
  \caption{Random forests and BART on MNIST ($n=10^4$, 10 classes)
  across the two-phase complexity axis $N^{\mathrm{max}}_{\mathrm{leaf}}/N_{\mathrm{tree}}$.
  \textbf{Top row}: clean training labels. \textbf{Bottom row}: $100\%$ random
  training labels. Both rows use clean test labels. Solid curves give the
  original single-forest train and test scores. The BART ensemble (green, dashed)
  averages regression outputs over 16 chains with 25 draws each, then clips
  and normalizes the ten scores for evaluation. Under both label settings the
  forest test cross-entropy (orange) exhibits a pronounced double-descent peak
  near the interpolation threshold. With random training labels, the BART
  ensemble cross-entropy stays near $\log 10$.}
  \label{fig:rf}
\end{figure}

\subsection{Random forests and BART}
\label{sec:exp-rf}

To compare with a structurally different model family, we now
consider random forests \citep{breiman2001random} on MNIST, following the
double-descent setup of \citet{belkin2019reconciling}. We train on
$n=10^4$ samples and sweep complexity along the two-phase axis
$N^{\mathrm{max}}_{\mathrm{leaf}}/N_{\mathrm{tree}}$: Phase~1 holds
$N_{\mathrm{tree}}=1$ and varies the maximum number of leaves
$N^{\mathrm{max}}_{\mathrm{leaf}}$ from small values up to the
interpolation threshold (where the train cross-entropy first reaches zero);
Phase~2 fixes $N^{\mathrm{max}}_{\mathrm{leaf}}$ at the interpolation
threshold and grows $N_{\mathrm{tree}}$. We run the experiment under both
clean labels and $100\%$ random labels. For comparison, we use a BART-style Gaussian regression ensemble
\citep{chipman2010bart}, averaging 16 chains with 25 retained draws each
before clipping and normalizing the ten class scores. This is an empirical
comparison, not a classification posterior predictive to which
\eqref{eq:gap-lower-bounds-III} directly applies.

Figure~\ref{fig:rf} shows train and test cross-entropy across the
complexity axis under both label settings. The forest test cross-entropy
exhibits a pronounced double-descent peak around the interpolation
threshold and then drops sharply once additional trees are added in
Phase~2; the peak is more dramatic under $100\%$ random labels. The
BART ensemble cross-entropy is substantially smaller near interpolation,
mirroring the attenuation of the peak in the CNN result.


\section{Conclusion}
\label{sec:conclusion}

We studied which component of the test loss is responsible for the
double-descent peak. We gave an upper bound on the MLE excess test cross-entropy
that decomposes into four interpretable terms: an approximation-and-shift
term, a minimax batch regret, a prior cost, and an MLE-vs-Bayes gap.
Under the posterior-sampling assumption, we also lower-bounded the
MLE-vs-Bayes gap by the expected MLE-vs-ensemble loss difference in the
well-specified setting. Through experiments under
fully random labels, where the approximation-and-shift term is zero by
construction, we found a large SGD-vs-ensemble gap near interpolation,
supporting the MLE-vs-Bayes gap as a major contributor to the peak. For practitioners
working in the interpolation regime, this picture suggests that the
double-descent peak can depend strongly on the prediction rule, and that
probability-averaged ensembles can attenuate it.

\textbf{Limitations.} Our experimental sweeps are less extensive than
prior empirical studies of double descent \citep{nakkiran2021deep}, do
not include the largest modern architectures, and use eight SGLD chains
per CNN width. The gap lower bound assumes an ideal posterior sampler.
Our finite SGLD runs may retain mixing and discretization error;
posterior consistency is assumed rather than proved for the CNNs studied here.
The longer-run and smaller-step checks support robustness of the observed
comparison, but do not certify the measured finite-test-set gap as a lower
bound. The universal compression decomposition relies
on a supremum over the model family in the batch-regret and MLE-vs-Bayes
terms, resulting in an upper bound rather than an equality; this is the
price for cleanly separating the batch regret from the MLE-vs-Bayes gap.
We do not establish numerical tightness for deep networks or directly
estimate (II), (II'), or the supremum in (III). Finally, while the
random-labels experiment eliminates (I) and shows a large empirical gap,
it does not establish that (III) dominates across all settings.

\begin{ack}
This work received no third-party funding or support.
The authors declare no competing interests.
\end{ack}

\small
\bibliographystyle{plainnat}
\bibliography{references}

\begin{thebibliography}{34}
\providecommand{\natexlab}[1]{#1}
\providecommand{\url}[1]{\texttt{#1}}
\expandafter\ifx\csname urlstyle\endcsname\relax
  \providecommand{\doi}[1]{doi: #1}\else
  \providecommand{\doi}{doi: \begingroup \urlstyle{rm}\Url}\fi

\bibitem[Advani and Saxe(2017)]{advani2017high}
Madhu~S. Advani and Andrew~M. Saxe.
\newblock High-dimensional dynamics of generalization error in neural networks.
\newblock \emph{arXiv preprint arXiv:1710.03667}, 2017.

\bibitem[Barron et~al.(1998)Barron, Rissanen, and Yu]{barron1998minimum}
Andrew Barron, Jorma Rissanen, and Bin Yu.
\newblock The minimum description length principle in coding and modeling.
\newblock \emph{IEEE Transactions on Information Theory}, 44\penalty0
  (6):\penalty0 2743--2760, 1998.

\bibitem[Bartlett and Mendelson(2002)]{bartlett2002rademacher}
Peter~L. Bartlett and Shahar Mendelson.
\newblock Rademacher and {G}aussian complexities: Risk bounds and structural
  results.
\newblock \emph{Journal of Machine Learning Research}, 3:\penalty0 463--482,
  2002.

\bibitem[Bartlett et~al.(2020)Bartlett, Long, Lugosi, and
  Tsigler]{bartlett2020benign}
Peter~L. Bartlett, Philip~M. Long, G{\'a}bor Lugosi, and Alexander Tsigler.
\newblock Benign overfitting in linear regression.
\newblock \emph{Proceedings of the National Academy of Sciences}, 117\penalty0
  (48):\penalty0 30063--30070, 2020.

\bibitem[Belkin et~al.(2019)Belkin, Hsu, Ma, and Mandal]{belkin2019reconciling}
Mikhail Belkin, Daniel Hsu, Siyuan Ma, and Soumik Mandal.
\newblock Reconciling modern machine-learning practice and the classical
  bias--variance trade-off.
\newblock \emph{Proceedings of the National Academy of Sciences}, 116\penalty0
  (32):\penalty0 15849--15854, 2019.

\bibitem[Belkin et~al.(2020)Belkin, Hsu, and Xu]{belkin2020two}
Mikhail Belkin, Daniel Hsu, and Ji~Xu.
\newblock Two models of double descent for weak features.
\newblock \emph{SIAM Journal on Mathematics of Data Science}, 2\penalty0
  (4):\penalty0 1167--1180, 2020.

\bibitem[Breiman(2001)]{breiman2001random}
Leo Breiman.
\newblock Random forests.
\newblock \emph{Machine Learning}, 45\penalty0 (1):\penalty0 5--32, 2001.

\bibitem[Chen et~al.(2021)Chen, Min, Belkin, and Karbasi]{chen2021multiple}
Lin Chen, Yifei Min, Mikhail Belkin, and Amin Karbasi.
\newblock Multiple descent: Design your own generalization curve.
\newblock In \emph{Advances in Neural Information Processing Systems}, 2021.

\bibitem[Chen et~al.(2023)Chen, Lau, Mendel, Wei, and
  Murfet]{chen2023dynamical}
Zhongtian Chen, Edmund Lau, Jake Mendel, Susan Wei, and Daniel Murfet.
\newblock Dynamical versus {B}ayesian phase transitions in a toy model of
  superposition.
\newblock \emph{arXiv preprint arXiv:2310.06301}, 2023.

\bibitem[Chipman et~al.(2010)Chipman, George, and McCulloch]{chipman2010bart}
Hugh~A. Chipman, Edward~I. George, and Robert~E. McCulloch.
\newblock {BART}: Bayesian additive regression trees.
\newblock \emph{The Annals of Applied Statistics}, 4\penalty0 (1):\penalty0
  266--298, 2010.
\newblock \doi{10.1214/09-AOAS285}.

\bibitem[Davisson(1973)]{davisson1973universal}
Lee~D. Davisson.
\newblock Universal noiseless coding.
\newblock \emph{IEEE Transactions on Information Theory}, 19\penalty0
  (6):\penalty0 783--795, 1973.

\bibitem[Hastie et~al.(2022)Hastie, Montanari, Rosset, and
  Tibshirani]{hastie2022surprises}
Trevor Hastie, Andrea Montanari, Saharon Rosset, and Ryan~J. Tibshirani.
\newblock Surprises in high-dimensional ridgeless least squares interpolation.
\newblock \emph{Annals of Statistics}, 50\penalty0 (2):\penalty0 949--986,
  2022.

\bibitem[Haussler and Opper(1997)]{haussler1997mutual}
David Haussler and Manfred Opper.
\newblock Mutual information, metric entropy and cumulative relative entropy
  risk.
\newblock \emph{Annals of Statistics}, 25\penalty0 (6):\penalty0 2451--2492,
  1997.

\bibitem[Hoogland et~al.(2024)Hoogland, Wang, Farrugia-Roberts, Carroll, Wei,
  and Murfet]{hoogland2024towards}
Jesse Hoogland, George Wang, Matthew Farrugia-Roberts, Liam Carroll, Susan Wei,
  and Daniel Murfet.
\newblock Loss landscape degeneracy and stagewise development in transformers.
\newblock \emph{arXiv preprint arXiv:2402.02364}, 2024.

\bibitem[Mei and Montanari(2022)]{mei2022generalization}
Song Mei and Andrea Montanari.
\newblock The generalization error of random features regression: Precise
  asymptotics and the double descent curve.
\newblock \emph{Communications on Pure and Applied Mathematics}, 75\penalty0
  (4):\penalty0 667--766, 2022.

\bibitem[Merhav and Feder(1998)]{merhav1998universal}
Neri Merhav and Meir Feder.
\newblock Universal prediction.
\newblock \emph{IEEE Transactions on Information Theory}, 44\penalty0
  (6):\penalty0 2124--2147, 1998.

\bibitem[Muthukumar et~al.(2020)Muthukumar, Vodrahalli, Subramanian, and
  Sahai]{muthukumar2020harmless}
Vidya Muthukumar, Kailas Vodrahalli, Vignesh Subramanian, and Anant Sahai.
\newblock Harmless interpolation of noisy data in regression.
\newblock \emph{IEEE Journal on Selected Areas in Information Theory},
  1\penalty0 (1):\penalty0 67--83, 2020.

\bibitem[Nakkiran(2019)]{nakkiran2020optimal}
Preetum Nakkiran.
\newblock More data can hurt for linear regression: Sample-wise double descent.
\newblock \emph{arXiv preprint arXiv:1912.07242}, 2019.

\bibitem[Nakkiran et~al.(2021{\natexlab{a}})Nakkiran, Kaplun, Bansal, Yang,
  Barak, and Sutskever]{nakkiran2021deep}
Preetum Nakkiran, Gal Kaplun, Yamini Bansal, Tristan Yang, Boaz Barak, and Ilya
  Sutskever.
\newblock Deep double descent: where bigger models and more data hurt.
\newblock \emph{Journal of Statistical Mechanics: Theory and Experiment},
  2021\penalty0 (12):\penalty0 124003, 2021{\natexlab{a}}.

\bibitem[Nakkiran et~al.(2021{\natexlab{b}})Nakkiran, Venkat, Kakade, and
  Ma]{nakkiran2021optimal}
Preetum Nakkiran, Prayaag Venkat, Sham~M. Kakade, and Tengyu Ma.
\newblock Optimal regularization can mitigate double descent.
\newblock In \emph{International Conference on Learning Representations},
  2021{\natexlab{b}}.

\bibitem[Polyanskiy and Wu(2024)]{polyanskiy-wu}
Yury Polyanskiy and Yihong Wu.
\newblock \emph{Information Theory: From Coding to Learning}.
\newblock Cambridge University Press, first edition, 2024.
\newblock Draft, August 2024.

\bibitem[Power et~al.(2022)Power, Burda, Edwards, Babuschkin, and
  Misra]{power2022grokking}
Alethea Power, Yuri Burda, Harri Edwards, Igor Babuschkin, and Vedant Misra.
\newblock Grokking: Generalization beyond overfitting on small algorithmic
  datasets.
\newblock \emph{arXiv preprint arXiv:2201.02177}, 2022.

\bibitem[Rissanen(1984)]{rissanen1984universal}
Jorma Rissanen.
\newblock Universal coding, information, prediction, and estimation.
\newblock \emph{IEEE Transactions on Information Theory}, 30\penalty0
  (4):\penalty0 629--636, 1984.

\bibitem[Rissanen(1996)]{rissanen1996fisher}
Jorma Rissanen.
\newblock Fisher information and stochastic complexity.
\newblock \emph{IEEE Transactions on Information Theory}, 42\penalty0
  (1):\penalty0 40--47, 1996.

\bibitem[Robert(2007)]{robert2007bayesian}
Christian~P. Robert.
\newblock \emph{The {B}ayesian Choice: From Decision-Theoretic Foundations to
  Computational Implementation}.
\newblock Springer, second edition, 2007.

\bibitem[Schaeffer et~al.(2023)Schaeffer, Khona, Robertson, Boopathy,
  Pistunova, Rocks, Fiete, and Koyejo]{schaeffer2023double}
Rylan Schaeffer, Mikail Khona, Zachary Robertson, Akhilan Boopathy, Kateryna
  Pistunova, Jason~W. Rocks, Ila~Rani Fiete, and Oluwasanmi Koyejo.
\newblock Double descent demystified: Identifying, interpreting \& ablating the
  sources of a deep learning puzzle.
\newblock \emph{arXiv preprint arXiv:2303.14151}, 2023.

\bibitem[Shtarkov(1987)]{shtarkov1987universal}
Yurii~M. Shtarkov.
\newblock Universal sequential coding of single messages.
\newblock \emph{Problems of Information Transmission}, 23\penalty0
  (3):\penalty0 3--17, 1987.

\bibitem[Steinke and Zakynthinou(2020)]{steinke2020reasoning}
Thomas Steinke and Lydia Zakynthinou.
\newblock Reasoning about generalization via conditional mutual information.
\newblock In \emph{Conference on Learning Theory}, 2020.

\bibitem[Teh et~al.(2016)Teh, Thiery, and Vollmer]{teh2016sgld}
Yee~Whye Teh, Alexandre~H. Thiery, and Sebastian~J. Vollmer.
\newblock Consistency and fluctuations for stochastic gradient langevin
  dynamics.
\newblock \emph{Journal of Machine Learning Research}, 17\penalty0
  (7):\penalty0 1--33, 2016.
\newblock URL \url{https://jmlr.org/papers/v17/teh16a.html}.

\bibitem[Vapnik and Chervonenkis(1971)]{vapnik1971uniform}
Vladimir~N. Vapnik and Alexey~Ya. Chervonenkis.
\newblock On the uniform convergence of relative frequencies of events to their
  probabilities.
\newblock \emph{Theory of Probability and Its Applications}, 16\penalty0
  (2):\penalty0 264--280, 1971.

\bibitem[Watanabe(2009)]{watanabe2009algebraic}
Sumio Watanabe.
\newblock \emph{Algebraic Geometry and Statistical Learning Theory}.
\newblock Cambridge University Press, 2009.

\bibitem[Welling and Teh(2011)]{welling2011sgld}
Max Welling and Yee~Whye Teh.
\newblock Bayesian learning via stochastic gradient langevin dynamics.
\newblock In \emph{Proceedings of the 28th International Conference on Machine
  Learning}, pages 681--688, 2011.
\newblock URL \url{https://icml.cc/2011/papers/398_icmlpaper.pdf}.

\bibitem[Wilson and Izmailov(2020)]{wilson2020bayesian}
Andrew~Gordon Wilson and Pavel Izmailov.
\newblock {B}ayesian deep learning and a probabilistic perspective of
  generalization.
\newblock In \emph{Advances in Neural Information Processing Systems}, 2020.

\bibitem[Xu and Raginsky(2017)]{xu2017information}
Aolin Xu and Maxim Raginsky.
\newblock Information-theoretic analysis of generalization capability of
  learning algorithms.
\newblock In \emph{Advances in Neural Information Processing Systems}, 2017.

\end{thebibliography}


\appendix



\section{Full training details}
\label{app:training}

\subsection{Shared CNN setup (SGD and SGLD comparison)}

\paragraph{Architecture.}
A ``$k$-width'' CNN: 4 convolutional blocks (channels $3 \to k \to 2k \to 4k
\to 8k$), each block consisting of a $3\times 3$ convolution (padding 1),
group normalization (one group), ReLU, and max pool with strides $[1, 2, 2, 8]$, so a
$32\times 32$ input collapses to $1\times 1$ after the four pools. A final
linear head maps $8k \to 10$. Models are trained from scratch with random
initialization; no pretraining and no data augmentation.

\paragraph{Dataset.}
CIFAR-10 with pixels rescaled to $[0,1]$ and standardized per-channel
using train-set mean and standard deviation. The first $n_{\mathrm{train}}
= 50{,}000$ images form the training set; the standard $10{,}000$-image
test images are unchanged. Labels are corrupted by replacing each label
with a uniformly random class with probability equal to the target noise
level, using a fixed RNG (\texttt{np.random.RandomState(seed=0)}) so that
the corruption pattern is identical across seeds and widths.

\paragraph{Optimization.}
SGD with no momentum and no weight decay; cross-entropy loss on the noisy
labels; fp32 throughout with TF32 disabled. Learning rate follows the
schedule
\[
\mathrm{lr}(t) \;=\; \mathrm{lr}_0 \,/\, (1 + t/T_{\mathrm{decay}})^{0.55},
\]
with $\mathrm{lr}_0 = 0.1$ and $T_{\mathrm{decay}} = 512$.
Minibatches of 128 examples are sampled uniformly with replacement.

\paragraph{Gaussian-prior SGLD.}
Every trainable parameter has an independent Gaussian prior. Convolutional
weights have mean zero and variance $2/\mathrm{fan\_in}$; linear weights
have mean zero and variance $1/\mathrm{fan\_in}$; biases have mean zero and
variance $0.25$; group-normalization scales have mean one and variance
$0.25$. For mean minibatch cross-entropy $\ell_B$, the update is
\[
\theta_{t+1}=\theta_t-\mathrm{lr}(t)
\left[\nabla\ell_B(\theta_t)-\frac{1}{n}\nabla\log\pi(\theta_t)\right]
+\sqrt{\frac{2\,\mathrm{lr}(t)}{n}}\,\xi_t,\qquad \xi_t\sim N(0,I).
\]
This is the temperature-one SGLD update \citep{welling2011sgld}, with
posterior target proportional to the likelihood times the stated prior.
Group normalization makes each prediction independent of the other
examples in its minibatch.

\paragraph{Per-seed protocol.}
Each seed receives its own \texttt{torch.manual\_seed(seed)} for weight
initialization and a CUDA \texttt{Generator(seed+123456)} for batch
sampling. Inputs and (corrupted) labels are identical across seeds; only
model initialization, minibatch order, and SGLD noise vary. For each (width $k$, seed)
pair we save the post-training softmax probabilities on the test set,
along with train cross-entropy and 0/1 error against the noisy training
labels, test cross-entropy against \emph{noisy} test labels (corrupted
with the same probability as the training labels), and test 0/1 error
against the \emph{clean} test labels. Test corruption uses
\texttt{np.random.RandomState(seed=12345)}, independently of training
corruption. SGLD discards the first $100{,}000$ steps and retains
predictive probabilities every $10{,}000$ steps thereafter, giving 10
draws per chain. Within-chain averages weight each retained draw by the
sum of step sizes since the preceding retained draw (or burn-in endpoint);
the final ensemble weights the eight chains equally. SGD predictions use
the final iterate. Log-loss evaluation floors assigned probabilities at
$10^{-12}$; the largest clipped fraction across widths is $0.504\%$ for
20\% noise and $0.223\%$ for 100\% noise, in the individual SGD predictions.

\paragraph{Test labels: clean vs.\ noisy.}
We use noisy test labels for cross-entropy and clean test labels for 0/1
error. For 0/1 error, expected noisy loss is an affine function of the
clean loss: writing the noise level as $\eta\in[0,1]$ and the number of
classes as $K=10$, for any classifier $\hat Y(X)$
\begin{equation}
\mathbb{E}_{\widetilde{Y}\mid Y}\!\left[\mathbb{1}\{\hat Y(X) \ne \widetilde{Y}\}\right]
\;=\; (1-\eta)\,\mathbb{1}\{\hat Y(X) \ne Y\}
\;+\; \eta\cdot\frac{K-1}{K},
\end{equation}
which is strictly-increasing affine in the clean 0/1 loss whenever
$\eta<1$, so the curve shape (peak and descent) is identical under either
choice up to an affine transformation. We use the clean 0/1 error for
interpretability since 0/1 is reported for reference rather than as a core
theoretical metric. For cross-entropy, by contrast, no such relation
holds:
\begin{equation}
\mathbb{E}_{\widetilde{Y}\mid Y}\!\left[-\log q(\widetilde{Y}\mid X)\right]
\;=\; (1-\eta)\,\bigl(-\log q(Y\mid X)\bigr)
\;-\; \frac{\eta}{K}\sum_{c=1}^{K} \log q(c\mid X),
\end{equation}
whose second term depends on the full predictive distribution rather
than only on $q(Y\mid X)$, so noisy cross-entropy is not a monotonic
function of clean cross-entropy. We therefore evaluate cross-entropy on
the noisy test labels, which are the relevant labels for the data
distribution $P^*$ in our decomposition.

\subsection{CIFAR-10 experiments}

\paragraph{20\% noise.}
Used in Figures~\ref{fig:ensemble-collapse} (Right), \ref{fig:ensemble}
(top row).
$n_{\mathrm{train}} = 50{,}000$, label noise $= 20\%$;
widths $k \in \{1, 2, 3, 4, 5, 6, 8, 10, 14, 20, 32, 48, 64\}$;
batch size $128$; $200{,}000$ gradient steps;
$K=8$ SGD seeds and $K=8$ SGLD chains per width;
\texttt{torch.compile} enabled.

\paragraph{100\% noise.}
Used in Figure~\ref{fig:ensemble} (bottom row).
$n_{\mathrm{train}} = 50{,}000$, label noise $= 100\%$ (all labels random);
widths $k \in \{1, 2, 3, 4, 5, 6, 8, 10, 14, 20, 32, 48, 64, 96, 128\}$.
All other settings as in the 20\% noise experiment.

\paragraph{SGD ensembling.}
Figure~\ref{fig:gap-vs-seeds} uses archived SGD predictions from 64 seeds
per width under 20\% noise and 16 under 100\% noise. The architecture is
as above with batch normalization in place of group normalization;
training and test inputs are standardized separately. SGD has no momentum
or weight decay and runs for $200{,}000$ steps with batch size 128 and
$\mathrm{lr}(t)=0.1/\sqrt{1+\lfloor t/512\rfloor}$.
Minibatches follow shuffled passes through the training set, with each
seed used for both initialization and the CUDA shuffling generator.
Widths are as above, with $\{192,256,384\}$ additionally included under
100\% noise. Label corruption and test metrics are as above. Ensembles
average the final-iterate class probabilities across seeds.

\subsection{Random forests on MNIST}

\paragraph{Dataset.}
MNIST with pixels divided by $255$ (range $[0,1]$); the official
$60{,}000$-train / $10{,}000$-test split is used. We uniformly subsample
$n = 10{,}000$ training images without replacement (RNG seed $42$) and
evaluate on the full $10{,}000$-image test set. Targets are one-hot
encoded over $K = 10$ classes.

\paragraph{Model.}
We use \texttt{sklearn.ensemble.RandomForestRegressor} with
\texttt{bootstrap=False} and \texttt{max\_features} $= \lfloor\sqrt{784}\rfloor = 28$.
The train and test curves use the archived single-forest results.

\paragraph{Two-phase complexity sweep.}
Along the $x$-axis $N^{\mathrm{max}}_{\mathrm{leaf}} / N_{\mathrm{tree}}$:
Phase~1 fixes $N_{\mathrm{tree}} = 1$ and varies
$N^{\mathrm{max}}_{\mathrm{leaf}}$ from small values up to $n=10^4$
(passing through the interpolation threshold); Phase~2 fixes
$N^{\mathrm{max}}_{\mathrm{leaf}}$ at the interpolation threshold and
varies $N_{\mathrm{tree}} \in \{2, 3, 5, 7, 10, 12, 15, 20\}$. We run two
versions: one with the clean MNIST labels (Phase~2 at
$N^{\mathrm{max}}_{\mathrm{leaf}}=2000$, just past the empirical
interpolation point), and one with $100\%$ random labels (Phase~2 at
$N^{\mathrm{max}}_{\mathrm{leaf}}=n=10{,}000$, since fully fitting random
labels requires one leaf per training point).

\paragraph{BART comparison.}
The comparison uses a sum of $m$ regression trees, each capped at $L$
leaves, with one-hot Gaussian regression variance $\sigma^2=0.1$ and
independent Gaussian leaf values of variance $0.0625/m$ and mean zero.
The split probability is $0.5$ at every depth, with uniform feature and
available-threshold proposals. Each of 16 chains uses 3,000 burn-in
sweeps and retains 25 draws, one every 120 sweeps; the number of structure
moves per tree per sweep is
$\max(1,\mathrm{round}((30L+3000)/3000))$.
The BART curve averages regression outputs across draws and chains, then
clips and normalizes the ten scores. This transformation is not the
categorical posterior predictive of the Gaussian regression model.

\paragraph{Metrics.}
We report squared loss (used only to identify the interpolation
threshold as the smallest $x$-value at which the average train squared
loss first falls below $10^{-4}$), zero-one loss, and cross-entropy.
Cross-entropy is computed on the row-renormalized, $\varepsilon$-clipped
one-hot-mean prediction with $\varepsilon = 10^{-12}$, ensuring a proper
probability vector.

\section{Compute resources}
\label{app:compute}

The CNN reruns used four AWS Spot instances, each with eight NVIDIA H100
80\,GB GPUs and a 200\,GB root volume. The 448 main runs and 48 sampling
checks used approximately 145 allocated H100 GPU-hours, including
provisioning and shutdown. Recorded training times were 6.9--62.6 minutes
per main run and 14.5--121.6 minutes per sampling check.
The archived forest and BART experiments ran on CPU; their hardware and
runtime details were not reconstructed. Hardware and runtime details for
the archived SGD runs in Figure~\ref{fig:gap-vs-seeds} were also not
reconstructed. These figures also exclude earlier exploratory runs.

\section{Notation and empirical quantities}
\label{app:interpretation}

\begin{center}
\begin{tabular}{lp{0.68\linewidth}}
\toprule
Quantity & Interpretation and measurement \\
\midrule
$L_n(P^*)$ & Expected plug-in test log-loss; estimated by the mean SGD test loss. \\
$Q^\pi_{n+1}$ & Bayes posterior predictive under the stated prior; approximated by SGLD. \\
(I) & Change in plug-in risk from distribution shift; zero under uniform random labels. \\
(II), (II') & Minimax risk and prior cost; not estimated in the CNN experiments. \\
(III) & Worst-case MLE-vs-Bayes risk difference; not directly estimated. \\
Empirical gap & SGD test loss minus SGLD ensemble test loss; the expectation lower bound requires unbiased posterior prediction. \\
\bottomrule
\end{tabular}
\end{center}

\section{Sampling checks and ensemble control}
\label{app:sampling}

We run four seeds at widths $\{8,14,64\}$ under 20\% noise and
$\{10,20,128\}$ under 100\% noise. The longer-run check uses 400,000
steps, discards 200,000, and retains a draw every 20,000 steps. The
smaller-step check uses 400,000 steps with
$\mathrm{lr}(t)=0.05/(1+t/1024)^{0.55}$, discards 200,000 steps, and
retains a draw every 20,000 steps, matching the main run's effective
integration time. Both comparisons use the same four seeds as their
200,000-step reference.

The largest absolute change in ensemble log-loss is at most $0.0186$ nats under
20\% noise and $0.000319$ nats under 100\% noise. Mean predictive total
variation from the reference ranges from $0.065$ to $0.117$ and from
$0.008$ to $0.018$, respectively. The qualitative loss comparison is
preserved, but predictive differences and retained-window drift remain;
these checks do not establish convergence.

\begin{figure}[t]
  \centering
  \includegraphics[width=\linewidth]{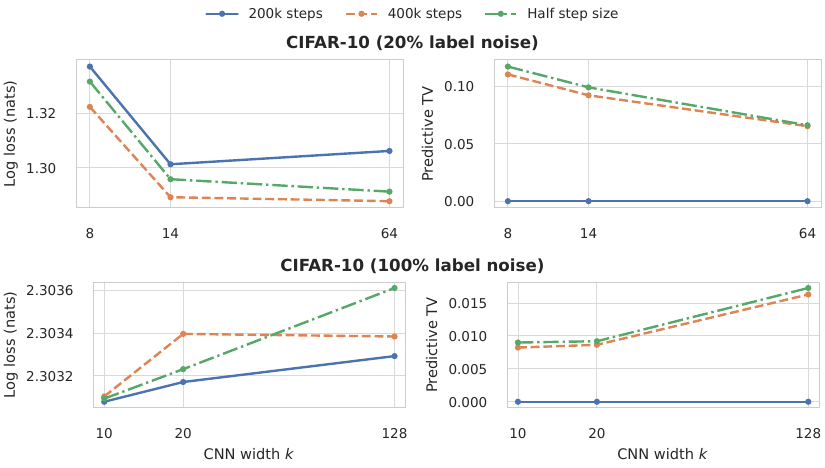}
  \caption{CIFAR-10 with 20\% label noise (top) and 100\% label noise
  (bottom). Each ensemble uses the same four seeds and 10 retained draws
  per chain. \textbf{Left:} Ensemble test cross-entropy for the main run,
  longer run, and smaller-step run. \textbf{Right:} Mean predictive total
  variation from the main-run ensemble at the same width.}
  \label{fig:sampling-checks}
\end{figure}

\begin{figure}[t]
  \centering
  \includegraphics[width=\linewidth]{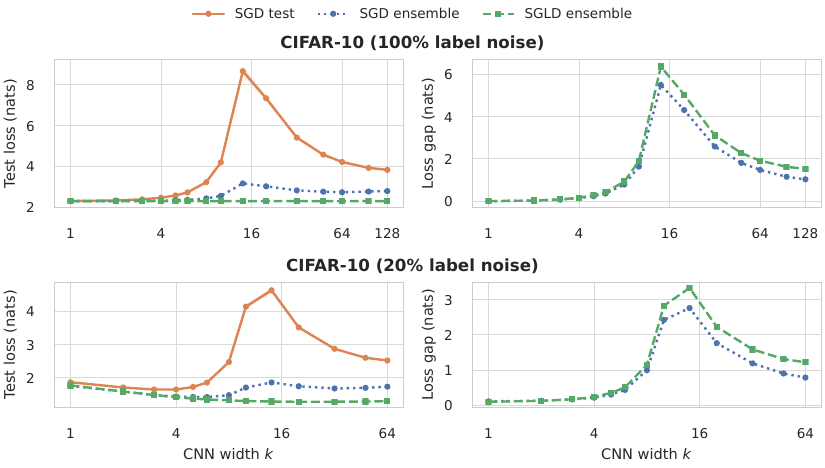}
  \caption{CIFAR-10 with 100\% label noise (top) and 20\% label noise
  (bottom), with $K=8$ seeds or chains per width.
  \textbf{Left:} Per-seed SGD test cross-entropy (orange; mean over seeds,
  band $\pm 1$ s.d.), the probability-averaged SGD ensemble (blue, dotted),
  and the SGLD ensemble (green, dashed).
  \textbf{Right:} The per-seed SGD mean minus each ensemble's test
  cross-entropy at the same width.}
  \label{fig:ensemble-control}
\end{figure}

\clearpage
\section*{NeurIPS Paper Checklist}

\begin{enumerate}

\item {\bf Claims}
    \item[] Question: Do the main claims made in the abstract and introduction accurately reflect the paper's contributions and scope?
    \item[] Answer: \answerYes{} 
    \item[] Justification: The abstract and Section~\ref{sec:intro} reflect the paper's contributions: a universal-compression decomposition for an arbitrary proper prior and experiments comparing SGD with Gaussian-prior SGLD under partial and fully random labels.
    \item[] Guidelines:
    \begin{itemize}
        \item The answer \answerNA{} means that the abstract and introduction do not include the claims made in the paper.
        \item The abstract and/or introduction should clearly state the claims made, including the contributions made in the paper and important assumptions and limitations. A \answerNo{} or \answerNA{} answer to this question will not be perceived well by the reviewers. 
        \item The claims made should match theoretical and experimental results, and reflect how much the results can be expected to generalize to other settings. 
        \item It is fine to include aspirational goals as motivation as long as it is clear that these goals are not attained by the paper. 
    \end{itemize}

\item {\bf Limitations}
    \item[] Question: Does the paper discuss the limitations of the work performed by the authors?
    \item[] Answer: \answerYes{} 
    \item[] Justification: We discuss finite-sampling error and the lack of a numerical tightness guarantee in Section~\ref{sec:conclusion}, and the empirical status of the forest/BART comparison in Section~\ref{sec:exp-rf}; ensemble-size and sampler sensitivity appear in Figure~\ref{fig:gap-vs-seeds} and Appendix~\ref{app:sampling}.
    \item[] Guidelines:
    \begin{itemize}
        \item The answer \answerNA{} means that the paper has no limitation while the answer \answerNo{} means that the paper has limitations, but those are not discussed in the paper. 
        \item The authors are encouraged to create a separate ``Limitations'' section in their paper.
        \item The paper should point out any strong assumptions and how robust the results are to violations of these assumptions (e.g., independence assumptions, noiseless settings, model well-specification, asymptotic approximations only holding locally). The authors should reflect on how these assumptions might be violated in practice and what the implications would be.
        \item The authors should reflect on the scope of the claims made, e.g., if the approach was only tested on a few datasets or with a few runs. In general, empirical results often depend on implicit assumptions, which should be articulated.
        \item The authors should reflect on the factors that influence the performance of the approach. For example, a facial recognition algorithm may perform poorly when image resolution is low or images are taken in low lighting. Or a speech-to-text system might not be used reliably to provide closed captions for online lectures because it fails to handle technical jargon.
        \item The authors should discuss the computational efficiency of the proposed algorithms and how they scale with dataset size.
        \item If applicable, the authors should discuss possible limitations of their approach to address problems of privacy and fairness.
        \item While the authors might fear that complete honesty about limitations might be used by reviewers as grounds for rejection, a worse outcome might be that reviewers discover limitations that aren't acknowledged in the paper. The authors should use their best judgment and recognize that individual actions in favor of transparency play an important role in developing norms that preserve the integrity of the community. Reviewers will be specifically instructed to not penalize honesty concerning limitations.
    \end{itemize}

\item {\bf Theory assumptions and proofs}
    \item[] Question: For each theoretical result, does the paper provide the full set of assumptions and a complete (and correct) proof?
    \item[] Answer: \answerYes{} 
    \item[] Justification: The main theoretical result---the test-loss decomposition in Equation~\ref{eq:decomp}---is stated with explicit assumptions (Section~\ref{sec:setup}) and proved in Section~\ref{sec:decomp}. All reference predictors and the MLE-vs-Bayes gap are defined in Section~\ref{sec:refobj}.
    \item[] Guidelines:
    \begin{itemize}
        \item The answer \answerNA{} means that the paper does not include theoretical results. 
        \item All the theorems, formulas, and proofs in the paper should be numbered and cross-referenced.
        \item All assumptions should be clearly stated or referenced in the statement of any theorems.
        \item The proofs can either appear in the main paper or the supplemental material, but if they appear in the supplemental material, the authors are encouraged to provide a short proof sketch to provide intuition. 
        \item Inversely, any informal proof provided in the core of the paper should be complemented by formal proofs provided in appendix or supplemental material.
        \item Theorems and Lemmas that the proof relies upon should be properly referenced. 
    \end{itemize}

    \item {\bf Experimental result reproducibility}
    \item[] Question: Does the paper fully disclose all the information needed to reproduce the main experimental results of the paper to the extent that it affects the main claims and/or conclusions of the paper (regardless of whether the code and data are provided or not)?
    \item[] Answer: \answerYes{} 
    \item[] Justification: Appendix~\ref{app:training} provides architecture, optimizer, prior, schedule, batch size, training steps, widths, and seed counts. Section~\ref{sec:exp-rf} and the appendix describe the forest/BART comparison.
    \item[] Guidelines:
    \begin{itemize}
        \item The answer \answerNA{} means that the paper does not include experiments.
        \item If the paper includes experiments, a \answerNo{} answer to this question will not be perceived well by the reviewers: Making the paper reproducible is important, regardless of whether the code and data are provided or not.
        \item If the contribution is a dataset and\slash or model, the authors should describe the steps taken to make their results reproducible or verifiable. 
        \item Depending on the contribution, reproducibility can be accomplished in various ways. For example, if the contribution is a novel architecture, describing the architecture fully might suffice, or if the contribution is a specific model and empirical evaluation, it may be necessary to either make it possible for others to replicate the model with the same dataset, or provide access to the model. In general. releasing code and data is often one good way to accomplish this, but reproducibility can also be provided via detailed instructions for how to replicate the results, access to a hosted model (e.g., in the case of a large language model), releasing of a model checkpoint, or other means that are appropriate to the research performed.
        \item While NeurIPS does not require releasing code, the conference does require all submissions to provide some reasonable avenue for reproducibility, which may depend on the nature of the contribution. For example
        \begin{enumerate}
            \item If the contribution is primarily a new algorithm, the paper should make it clear how to reproduce that algorithm.
            \item If the contribution is primarily a new model architecture, the paper should describe the architecture clearly and fully.
            \item If the contribution is a new model (e.g., a large language model), then there should either be a way to access this model for reproducing the results or a way to reproduce the model (e.g., with an open-source dataset or instructions for how to construct the dataset).
            \item We recognize that reproducibility may be tricky in some cases, in which case authors are welcome to describe the particular way they provide for reproducibility. In the case of closed-source models, it may be that access to the model is limited in some way (e.g., to registered users), but it should be possible for other researchers to have some path to reproducing or verifying the results.
        \end{enumerate}
    \end{itemize}

\item {\bf Open access to data and code}
    \item[] Question: Does the paper provide open access to the data and code, with sufficient instructions to faithfully reproduce the main experimental results, as described in supplemental material?
    \item[] Answer: \answerYes{} 
    \item[] Justification: The complete code, data preparation scripts, and reproduction instructions are provided as supplementary material. The experiments use the publicly available CIFAR-10 and MNIST datasets, both downloaded automatically by the scripts; full hyperparameters and run-level details are in Appendix~\ref{app:training}.
    \item[] Guidelines:
    \begin{itemize}
        \item The answer \answerNA{} means that paper does not include experiments requiring code.
        \item Please see the NeurIPS code and data submission guidelines (\url{https://neurips.cc/public/guides/CodeSubmissionPolicy}) for more details.
        \item While we encourage the release of code and data, we understand that this might not be possible, so \answerNo{} is an acceptable answer. Papers cannot be rejected simply for not including code, unless this is central to the contribution (e.g., for a new open-source benchmark).
        \item The instructions should contain the exact command and environment needed to run to reproduce the results. See the NeurIPS code and data submission guidelines (\url{https://neurips.cc/public/guides/CodeSubmissionPolicy}) for more details.
        \item The authors should provide instructions on data access and preparation, including how to access the raw data, preprocessed data, intermediate data, and generated data, etc.
        \item The authors should provide scripts to reproduce all experimental results for the new proposed method and baselines. If only a subset of experiments are reproducible, they should state which ones are omitted from the script and why.
        \item At submission time, to preserve anonymity, the authors should release anonymized versions (if applicable).
        \item Providing as much information as possible in supplemental material (appended to the paper) is recommended, but including URLs to data and code is permitted.
    \end{itemize}

\item {\bf Experimental setting/details}
    \item[] Question: Does the paper specify all the training and test details (e.g., data splits, hyperparameters, how they were chosen, type of optimizer) necessary to understand the results?
    \item[] Answer: \answerYes{} 
    \item[] Justification: Appendix~\ref{app:training} specifies the SGD and Gaussian-prior SGLD protocols, including the batch size (128), learning-rate schedule, 200,000 main-run steps, widths, seed counts, and 50,000 CIFAR-10 training examples. Appendix~\ref{app:sampling} specifies the additional checks.
    \item[] Guidelines:
    \begin{itemize}
        \item The answer \answerNA{} means that the paper does not include experiments.
        \item The experimental setting should be presented in the core of the paper to a level of detail that is necessary to appreciate the results and make sense of them.
        \item The full details can be provided either with the code, in appendix, or as supplemental material.
    \end{itemize}

\item {\bf Experiment statistical significance}
    \item[] Question: Does the paper report error bars suitably and correctly defined or other appropriate information about the statistical significance of the experiments?
    \item[] Answer: \answerYes{} 
    \item[] Justification: The per-seed test cross-entropy curves report the mean over $K$ independently initialized seeds with a $\pm 1$ standard-deviation band across seeds (Figure~\ref{fig:ensemble}); sensitivity to the number of seeds is analyzed separately (Figure~\ref{fig:gap-vs-seeds}).
    \item[] Guidelines:
    \begin{itemize}
        \item The answer \answerNA{} means that the paper does not include experiments.
        \item The authors should answer \answerYes{} if the results are accompanied by error bars, confidence intervals, or statistical significance tests, at least for the experiments that support the main claims of the paper.
        \item The factors of variability that the error bars are capturing should be clearly stated (for example, train/test split, initialization, random drawing of some parameter, or overall run with given experimental conditions).
        \item The method for calculating the error bars should be explained (closed form formula, call to a library function, bootstrap, etc.)
        \item The assumptions made should be given (e.g., Normally distributed errors).
        \item It should be clear whether the error bar is the standard deviation or the standard error of the mean.
        \item It is OK to report 1-sigma error bars, but one should state it. The authors should preferably report a 2-sigma error bar than state that they have a 96\% CI, if the hypothesis of Normality of errors is not verified.
        \item For asymmetric distributions, the authors should be careful not to show in tables or figures symmetric error bars that would yield results that are out of range (e.g., negative error rates).
        \item If error bars are reported in tables or plots, the authors should explain in the text how they were calculated and reference the corresponding figures or tables in the text.
    \end{itemize}

\item {\bf Experiments compute resources}
    \item[] Question: For each experiment, does the paper provide sufficient information on the computer resources (type of compute workers, memory, time of execution) needed to reproduce the experiments?
    \item[] Answer: \answerNo{}
    \item[] Justification: Appendix~\ref{app:compute} reports the CNN rerun hardware, approximately 145 allocated H100 GPU-hours, and per-run training times. Hardware and runtime details for the archived SGD and CPU experiments were not reconstructed, and the total excludes earlier exploratory runs.
    \item[] Guidelines:
    \begin{itemize}
        \item The answer \answerNA{} means that the paper does not include experiments.
        \item The paper should indicate the type of compute workers CPU or GPU, internal cluster, or cloud provider, including relevant memory and storage.
        \item The paper should provide the amount of compute required for each of the individual experimental runs as well as estimate the total compute. 
        \item The paper should disclose whether the full research project required more compute than the experiments reported in the paper (e.g., preliminary or failed experiments that didn't make it into the paper). 
    \end{itemize}
    
\item {\bf Code of ethics}
    \item[] Question: Does the research conducted in the paper conform, in every respect, with the NeurIPS Code of Ethics \url{https://neurips.cc/public/EthicsGuidelines}?
    \item[] Answer: \answerYes{} 
    \item[] Justification: The work uses publicly available datasets and standard ML libraries, and conforms with the NeurIPS Code of Ethics.
    \item[] Guidelines:
    \begin{itemize}
        \item The answer \answerNA{} means that the authors have not reviewed the NeurIPS Code of Ethics.
        \item If the authors answer \answerNo, they should explain the special circumstances that require a deviation from the Code of Ethics.
        \item The authors should make sure to preserve anonymity (e.g., if there is a special consideration due to laws or regulations in their jurisdiction).
    \end{itemize}

\item {\bf Broader impacts}
    \item[] Question: Does the paper discuss both potential positive societal impacts and negative societal impacts of the work performed?
    \item[] Answer: \answerNA{} 
    \item[] Justification: This is foundational theoretical work on the test-loss behavior of trained models; we do not anticipate direct positive or negative societal impacts beyond contributing to the theoretical understanding of deep learning generalization.
    \item[] Guidelines:
    \begin{itemize}
        \item The answer \answerNA{} means that there is no societal impact of the work performed.
        \item If the authors answer \answerNA{} or \answerNo, they should explain why their work has no societal impact or why the paper does not address societal impact.
        \item Examples of negative societal impacts include potential malicious or unintended uses (e.g., disinformation, generating fake profiles, surveillance), fairness considerations (e.g., deployment of technologies that could make decisions that unfairly impact specific groups), privacy considerations, and security considerations.
        \item The conference expects that many papers will be foundational research and not tied to particular applications, let alone deployments. However, if there is a direct path to any negative applications, the authors should point it out. For example, it is legitimate to point out that an improvement in the quality of generative models could be used to generate Deepfakes for disinformation. On the other hand, it is not needed to point out that a generic algorithm for optimizing neural networks could enable people to train models that generate Deepfakes faster.
        \item The authors should consider possible harms that could arise when the technology is being used as intended and functioning correctly, harms that could arise when the technology is being used as intended but gives incorrect results, and harms following from (intentional or unintentional) misuse of the technology.
        \item If there are negative societal impacts, the authors could also discuss possible mitigation strategies (e.g., gated release of models, providing defenses in addition to attacks, mechanisms for monitoring misuse, mechanisms to monitor how a system learns from feedback over time, improving the efficiency and accessibility of ML).
    \end{itemize}
    
\item {\bf Safeguards}
    \item[] Question: Does the paper describe safeguards that have been put in place for responsible release of data or models that have a high risk for misuse (e.g., pre-trained language models, image generators, or scraped datasets)?
    \item[] Answer: \answerNA{} 
    \item[] Justification: The paper does not release pre-trained models or scraped datasets that pose misuse risks.
    \item[] Guidelines:
    \begin{itemize}
        \item The answer \answerNA{} means that the paper poses no such risks.
        \item Released models that have a high risk for misuse or dual-use should be released with necessary safeguards to allow for controlled use of the model, for example by requiring that users adhere to usage guidelines or restrictions to access the model or implementing safety filters. 
        \item Datasets that have been scraped from the Internet could pose safety risks. The authors should describe how they avoided releasing unsafe images.
        \item We recognize that providing effective safeguards is challenging, and many papers do not require this, but we encourage authors to take this into account and make a best faith effort.
    \end{itemize}

\item {\bf Licenses for existing assets}
    \item[] Question: Are the creators or original owners of assets (e.g., code, data, models), used in the paper, properly credited and are the license and terms of use explicitly mentioned and properly respected?
    \item[] Answer: \answerYes{} 
    \item[] Justification: We use the publicly available CIFAR-10 and MNIST datasets and standard libraries (PyTorch, scikit-learn); all assets are used under their respective licenses and standard citations are provided.
    \item[] Guidelines:
    \begin{itemize}
        \item The answer \answerNA{} means that the paper does not use existing assets.
        \item The authors should cite the original paper that produced the code package or dataset.
        \item The authors should state which version of the asset is used and, if possible, include a URL.
        \item The name of the license (e.g., CC-BY 4.0) should be included for each asset.
        \item For scraped data from a particular source (e.g., website), the copyright and terms of service of that source should be provided.
        \item If assets are released, the license, copyright information, and terms of use in the package should be provided. For popular datasets, \url{paperswithcode.com/datasets} has curated licenses for some datasets. Their licensing guide can help determine the license of a dataset.
        \item For existing datasets that are re-packaged, both the original license and the license of the derived asset (if it has changed) should be provided.
        \item If this information is not available online, the authors are encouraged to reach out to the asset's creators.
    \end{itemize}

\item {\bf New assets}
    \item[] Question: Are new assets introduced in the paper well documented and is the documentation provided alongside the assets?
    \item[] Answer: \answerNA{} 
    \item[] Justification: The paper does not release new datasets, pretrained models, or other assets.
    \item[] Guidelines:
    \begin{itemize}
        \item The answer \answerNA{} means that the paper does not release new assets.
        \item Researchers should communicate the details of the dataset\slash code\slash model as part of their submissions via structured templates. This includes details about training, license, limitations, etc. 
        \item The paper should discuss whether and how consent was obtained from people whose asset is used.
        \item At submission time, remember to anonymize your assets (if applicable). You can either create an anonymized URL or include an anonymized zip file.
    \end{itemize}

\item {\bf Crowdsourcing and research with human subjects}
    \item[] Question: For crowdsourcing experiments and research with human subjects, does the paper include the full text of instructions given to participants and screenshots, if applicable, as well as details about compensation (if any)? 
    \item[] Answer: \answerNA{} 
    \item[] Justification: The paper does not involve crowdsourcing or research with human subjects.
    \item[] Guidelines:
    \begin{itemize}
        \item The answer \answerNA{} means that the paper does not involve crowdsourcing nor research with human subjects.
        \item Including this information in the supplemental material is fine, but if the main contribution of the paper involves human subjects, then as much detail as possible should be included in the main paper. 
        \item According to the NeurIPS Code of Ethics, workers involved in data collection, curation, or other labor should be paid at least the minimum wage in the country of the data collector. 
    \end{itemize}

\item {\bf Institutional review board (IRB) approvals or equivalent for research with human subjects}
    \item[] Question: Does the paper describe potential risks incurred by study participants, whether such risks were disclosed to the subjects, and whether Institutional Review Board (IRB) approvals (or an equivalent approval/review based on the requirements of your country or institution) were obtained?
    \item[] Answer: \answerNA{} 
    \item[] Justification: The paper does not involve research with human subjects.
    \item[] Guidelines:
    \begin{itemize}
        \item The answer \answerNA{} means that the paper does not involve crowdsourcing nor research with human subjects.
        \item Depending on the country in which research is conducted, IRB approval (or equivalent) may be required for any human subjects research. If you obtained IRB approval, you should clearly state this in the paper. 
        \item We recognize that the procedures for this may vary significantly between institutions and locations, and we expect authors to adhere to the NeurIPS Code of Ethics and the guidelines for their institution. 
        \item For initial submissions, do not include any information that would break anonymity (if applicable), such as the institution conducting the review.
    \end{itemize}

\item {\bf Declaration of LLM usage}
    \item[] Question: Does the paper describe the usage of LLMs if it is an important, original, or non-standard component of the core methods in this research? Note that if the LLM is used only for writing, editing, or formatting purposes and does \emph{not} impact the core methodology, scientific rigor, or originality of the research, declaration is not required.
    \item[] Answer: \answerNA{}
    \item[] Justification: LLMs are not components of the evaluated learning algorithms. LLM assistance was used for coding and manuscript preparation.
    \item[] Guidelines:
    \begin{itemize}
        \item The answer \answerNA{} means that the core method development in this research does not involve LLMs as any important, original, or non-standard components.
        \item Please refer to our LLM policy in the NeurIPS handbook for what should or should not be described.
    \end{itemize}

\end{enumerate}

\end{document}